\documentclass{bmvc2k}

\usepackage{times}
\usepackage{epsfig}
\usepackage{graphicx}
\usepackage{amsmath}
\usepackage{amssymb}
\usepackage{calrsfs}
\usepackage{array}
\usepackage{booktabs}
\usepackage{multirow}
\usepackage{multicol}
\usepackage{wrapfig}
\usepackage[most]{tcolorbox}

\makeatletter
\DeclareRobustCommand\onedot{\futurelet\@let@token\@onedot}
\def\@onedot{\ifx\@let@token.\else.\null\fi\xspace}

\def\ie{\emph{i.e}\onedot}

\def\etal{\emph{et al}\onedot}
\makeatother

\DeclareMathAlphabet{\pazocal}{OMS}{zplm}{m}{n}

\newcommand{\best}[1]{\color{red}\textbf{#1}}
\newcommand{\secondb}[1]{\color{blue}\textbf{#1}}

\title{Subpixel Heatmap Regression for Facial Landmark Localization}

\addauthor{Adrian Bulat}{adrian@adrianbulat.com}{1}
\addauthor{Enrique Sanchez}{e.lozano@samsung.com}{1}
\addauthor{Georgios Tzimiropoulos}{g.tzimiropoulos@qmul.ac.uk}{1,2}

\addinstitution{
 Samsung AI Center\\
 Cambridge, UK
}
\addinstitution{
 Queen Mary University London\\
 London, UK
}

\runninghead{Bulat et al.}{Subpixel Heatmap regres. for Facial Landmark Localization}

\def\etal{\emph{et al}\bmvaOneDot}

\begin{document}

\maketitle

\begin{abstract}
   
   Deep Learning models based on heatmap regression have revolutionized the task of facial landmark localization with existing models working robustly under large poses, non-uniform illumination and shadows, occlusions and self-occlusions, low resolution and blur. However, despite their wide adoption, heatmap regression approaches suffer from discretization-induced errors related to both the heatmap encoding and decoding process. In this work we show that these errors have a surprisingly large negative impact on facial alignment accuracy. To alleviate this problem, we propose a new approach for the heatmap encoding and decoding process by leveraging the underlying continuous distribution. To take full advantage of the newly proposed encoding-decoding mechanism, we also introduce a Siamese-based training that enforces heatmap consistency across various geometric image transformations. Our approach offers noticeable gains across multiple datasets setting a new state-of-the-art result in facial landmark localization. Code alongside the pretrained models will be made available \href{https://www.adrianbulat.com/face-alignment}{here}.
   
\end{abstract}

\section{Introduction}

This paper is on the popular task of localizing landmarks (or keypoints) on the human face, also known as facial landmark localization or face alignment. Current state-of-the-art is represented by fully convolutional networks trained to perform heatmap regression~\cite{bulat2017far,sun2019deep,kumar2020luvli,tang2019towards,feng2018wing,wang2020deep}. Such methods can work robustly under large poses, non-uniform illumination and shadows, occlusions and self-occlusions~\cite{bulat2016two,bulat2017far,sun2019high,kumar2020luvli} and even very low resolution~\cite{bulat2018super}. However, despite their wide adoption, heatmap-based regression approaches suffer from discretization-induced errors. Although this is in general known, there are very few papers that study this problem~\cite{wan2020robust, tai2019towards, luvizon20182d}. Yet, in this paper, we show that this overlooked problem makes actually has surprisingly negative impact on the accuracy of the model. 

In particular, as working in high resolutions is computationally and memory prohibitive, typically, heatmap regression networks make predictions at $\frac{1}{4}$ of the input resolution~\cite{bulat2017far}. Note that the input image may already be a downsampled version of the original facial image. Due to the heatmap construction process that discretizes all values into a grid and the subsequent estimation process that consists of finding the coordinates of the maximum, large discretization errors are introduced. This in turn causes at least two problems: (a) the encoding process forces the network to learn randomly displaced points and, (b) the inference process of the decoder is done on a discrete grid failing to account for the continuous underlying Gaussian distribution of the heatmap. 

To alleviate the above problem, in this paper, we make the following \textbf{contributions}:

\begin{itemize}
    \item We rigorously study and propose a continuous method for heatmap regression, consisting of a simple continuous heatmap encoding and a newly proposed continuous heatmap decoding method, called local-softargmax, that largely solve the quantization errors introduced by the heatmap discretization process.
    \item We also propose an accompanying Siamese-based training procedure that enforces consistent heatmap predictions across various geometric image transformations. 
    \item 
     By largely alleviating the quantization problem with the proposed solutions, we show that the standard method of~\cite{bulat2017far} sets a new state-of-the-art on multiple datasets, offering significant improvements over prior-work.
\end{itemize}

\section{Related work}

Most recent efforts on improving the accuracy of face alignment fall into one of the following two categories: network architecture improvements and loss function improvements.

\noindent\textbf{Network architectural improvements:} The first work to popularize and make use of encoder-decoder models with heatmap-based regression for face alignment was the work of Bulat\&Tzimiropoulos~\cite{bulat2017far} where the authors adapted an HourGlass network~\cite{newell2016stacked} with 4 stages and the Hierarchical Block of~\cite{bulat2017binarized} for face alignment. Subsequent works generally preserved the same style of U-Net~\cite{ronneberger2015u} and Hourglass structures with notable differences in~\cite{xiao2018simple,sun2019high,wang2020deep} where the authors used ResNets~\cite{he2016deep} adapted for dense pixel-wise predictions. More specifically, in~\cite{xiao2018simple}, the authors removed the last fully connected layer and the global pooling operation from a ResNet model and then attempted to recover the lost resolution using  a series of convolutions and deconvolutional layers. In~\cite{wang2020deep}, Wang~\etal expanded upon this by introducing a novel structure that connects high-to-low  convolution streams in  parallel, maintaining the  high-resolution representations through  the entire model. Building on top of~\cite{bulat2017far}, in CU-Net~\cite{tang2019towards} and DU-Net~\cite{tang2018quantized} Tang~\etal combined U-Nets with DenseNet-like~\cite{huang2017densely} architectures connecting the $i$-th U-Net with all previous ones via skip connections.

\noindent\textbf{Loss function improvements:} The standard loss typically used for heatmap regression is a pixel-wise $\ell_2$ or $\ell_1$ loss~\cite{bulat2016convolutional,bulat2016two,bulat2017far,sun2019deep,wang2020deep,tang2019towards}. Feng~\etal~\cite{feng2018wing} argued that more attention should be payed to small and medium range errors during training, introducing the Wing loss that amplifies the impact of the errors within a defined interval by switching from an $\ell_1$ to a modified $\log$-based loss. Improving upon this, in~\cite{wang2019adaptive}, the authors introduced the Adaptive Wing Loss, a loss capable to update its curvature based on the ground truth pixels. The predictions are further aided by the integration of coordinates encoding via CoordConv~\cite{liu2018intriguing} into the model.
In~\cite{kumar2020luvli}, Kumar~\etal introduced the so-called LUVLi loss that jointly optimizes the location of the keypoints, the uncertainty, and the visibility likelihood. Albeit for human pose estimation,~\cite{luvizon20182d} proposes an alternative to heatmap-based regression by introducing a differential soft-argmax function applied globally to the output features. However, the lack of structure induced by a Gaussian prior,  hinders their accuracy.

Contrary to the aforementioned works, we attempt to address the quantization-induced error by proposing a simple continuous approach to the heatmap encoding and decoding process. In this direction,~\cite{tai2019towards} proposes an analytic solution to obtain the fractional shift by assuming that the generated heatmap follows a Gaussian distribution and applies this to stabilize facial landmark localization in video. A similar assumption is made by~\cite{wan2020robust} which solves an optimization problem to obtain the subpixel solution. Finally, \cite{luvizon20182d} uses global softargmax. Our method is mostly similar to~\cite{luvizon20182d} which we compare with in Section~\ref{sec:ablation}.


\section{Method}\label{sec:method}

\subsection{Preliminaries}\label{ssec:preliminaries}

Given a training sample $(\mathbf{X}, \mathbf{y})$, with $\mathbf{y}\in \mathbb{R}^{k\times 2}$ denoting the coordinates of the $K$ joints in the corresponding image $\mathbf{X}$, current facial landmark localization methods encode the target ground truth coordinates as a set of $k$ heatmaps with a 2D Gaussian centered at them:
\begin{equation} \label{eq:heatmap}
    \pazocal{G}_{i,j,k}(\mathbf{y}) = \frac{1}{2\pi \sigma^2}e^{-\frac{1}{2\sigma^2}[(i-\tilde{y}^{[1]}_k)^2+(j-\tilde{y}^{[2]}_k)^2]},
\end{equation}
where $y^{[1]}_k$ and $y^{[2]}_k$ are the spatial coordinates of the $k$-th point, and $\tilde{y}^{[1]}_k$ and $\tilde{y}^{[2]}_k$ their scaled, quantized version:

\begin{equation}
    \label{eq:approx}
    (\tilde{y}^{[1]}_k, \tilde{y}^{[2]}_k) = (\lfloor \frac{1}{s} y^{[1]}_k \rceil, \lfloor \frac{1}{s} y^{[2]}_k \rceil)
\end{equation}
where $\lfloor . \rceil$ is the rounding operator and $1/s$ is the scaling factor used to scale the image to a pre-defined resolution. $\sigma$ is the variance, a fixed value which is task and dataset dependent. For a given set of landmarks $\mathbf{y}$, Eq.~\ref{eq:heatmap} produces a corresponding heatmap $\pazocal{H}\in\mathbb{R}^{k\times W_{hm}\times H_{hm}}$.

Heatmap-based regression overcomes the lack of a spatial and contextual information of direct coordinate regression. Not only such representations are easier to learn by allowing visually similar parts to produce proportionally high responses instead of predicting a unique value, but they are also more interpretable and semantically meaningful.

\subsection{Continuous Heatmap Encoding}\label{ssec:heatmap-encoding}

Despite the advantages of heatmap regression, one key inherent issue with the approach has been overlooked: The heatmap generation process introduces relatively high quantization errors. This is a direct consequence of the trade-offs made during the generation process: since generating the heatmaps predictions at the original image resolution is prohibitive, the localization process involves cropping and re-scaling the facial images such that the final predicted heatmaps are typically at a $64\times64$px resolution~\cite{bulat2017far}. 
As described in Section~\ref{ssec:preliminaries}, this process re-scales and quantizes the landmark coordinates as
$\hat{\mathbf{y}} = \texttt{quantize}(\frac{1}{s}\mathbf{y})$, where \texttt{round} or \texttt{floor} is the quantization function. However, there is no need to quantize. One can simply create a Gaussian located at:
\begin{equation}
    \label{eq:approx-no-quant}
    (\tilde{y}^{[1]}_k, \tilde{y}^{[2]}_k) = ( \frac{1}{s} y^{[1]}_k, \frac{1}{s} y^{[2]}_k),
\end{equation}
and then sample it over a regular spatial grid. This will completely remove the quantization error introduced previously and will only add some aliasing due to the sampling process.

\subsection{Continuous Heatmap Decoding with Local Soft-argmax}\label{ssec:heatmap-decoding}

Currently, the typical landmark localization process from 2D heatmaps consists of finding the location of the pixel with the highest value~\cite{bulat2017far}. This is typically followed by a heuristic correction with $0.25$px toward the location of the second highest neighboring pixel. The goal of this adjustment is to partially compensate for the effect induced by the quantization process: on one side by the heatmap generation process itself (as described in Section~\ref{ssec:heatmap-encoding}) and on other side, by the coarse nature of the predicted heatmap that uses the maximum value solely as the location of the point. We note that, despite the fact that the ground truth heatmaps are affected by quantization errors, generally, the networks learns to adjust, to some extent its predictions, making the later heuristic correction work well in practice.

Rather than using the above heuristic, we propose to predict the location of the keypoint by analyzing the pixels in its neighbourhood and exploiting the known targeted Gaussian distribution. For a given heatmap $\pazocal{H}_k$, we firstly find the coordinates corresponding to the maximum value $(\hat{y}_k^{[1]}, \hat{y}_k^{[2]}) = \arg \max{\pazocal{H}_k}$ and then, around this location, we select a small square matrix $h_k$ of size $d\times d$, where $l=\frac{d}{2}$. Then, we predict an offset $(\Delta\hat{y}_k^{[1]},\Delta\hat{y}_k^{[2]})$ by finding a soft continuous maximum value within the selected matrix, effectively retrieving a correction, using a local soft-argmax:
\begin{equation}
    (\Delta\hat{y}_k^{[1]},\Delta\hat{y}_k^{[2]}) = \sum_{m,n}\texttt{softmax}(\tau h_k)_{m,n}(m,n),
\end{equation}
where $\tau$ is the temperature that controls the resulting probability map, and $(m,n)$ are the indices that iterate over the pixel coordinates of the heatmap $h_k$. $\texttt{softmax}$ is defined as: 
\begin{equation}
    \texttt{softmax}(h)_{m,n} = \frac{e^{ h_{m,n}}}{\sum_{m',n'} e^{ h_{m',n'}}}
\end{equation}

\begin{figure}
    \centering
    \includegraphics[height=3.5cm,trim=0.5cm 3.5cm 0.5cm 2.5cm, clip]{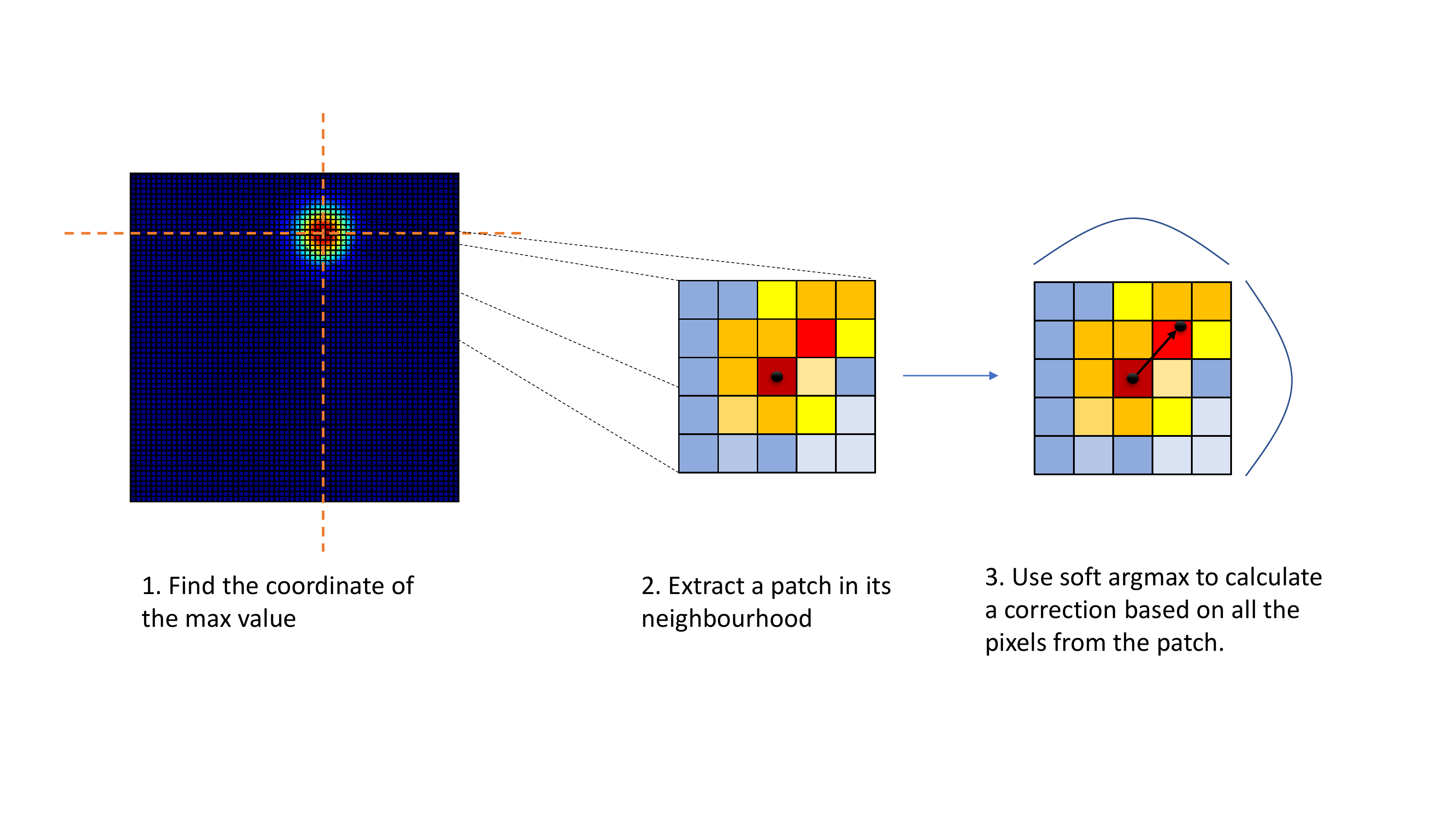}
    \caption{\textbf{Proposed heatmap decoding.} Given a predicted heatmap, (1) we find the location of the maximum, (2) and then crop around it a $k\times k$ patch. Finally, (3) we apply a soft-argmax on the patch and retrieve a correction applied to the location estimated at step (1).}
    \label{fig:heatmap-decoding}
    \vspace{-10px}
\end{figure}

The final prediction is then obtained as:
    $(\hat{y}_k^{[1]} + \Delta\hat{y}_k^{[1]}-l, \hat{y}_k^{[2]} + \Delta\hat{y}_k^{[2]} -l).$
The 3 step process is illustrated in Fig.~\ref{fig:heatmap-decoding}.

\subsection{Siamese consistency training}\label{ssec:siamese}


\begin{figure}[ht!]
    \centering
    \includegraphics[height=2.8cm,trim=0.1cm 0.1cm 0.1cm 0.1cm, clip]{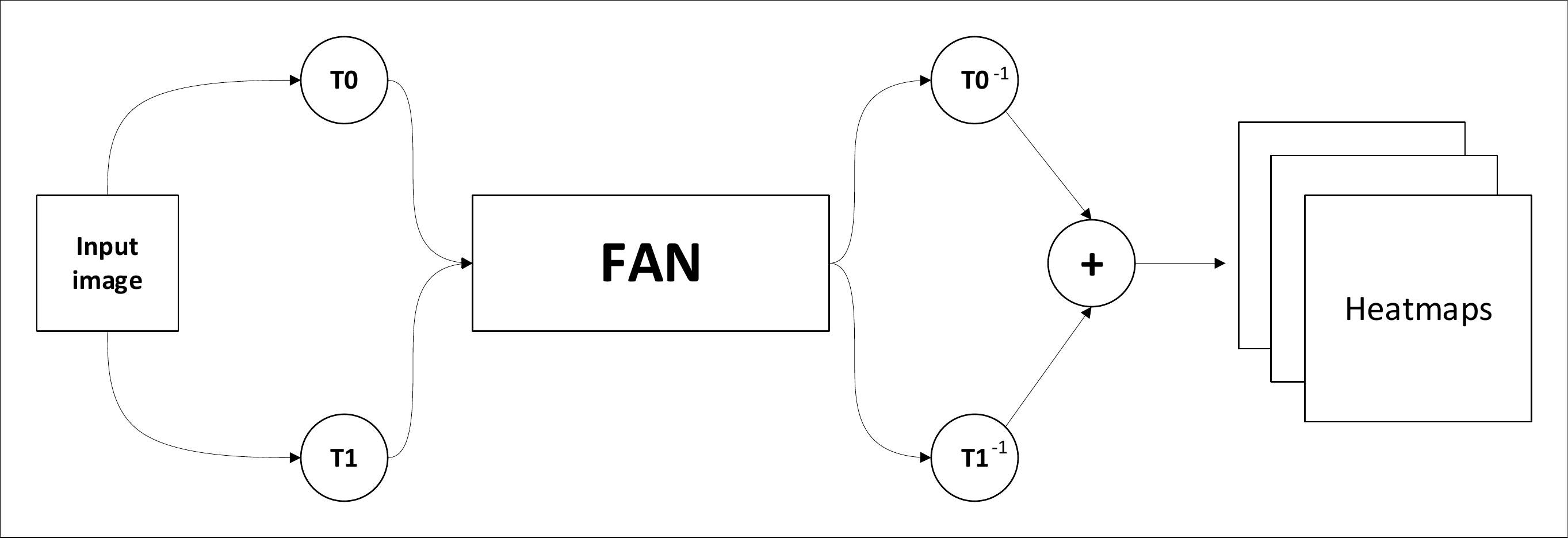}
    \caption{\textbf{Siamese transformation-invariant training.} T0 and T1 are two randomly sampled data augmentation transformations applied on the input image. After passing the augmented images through the network a set of heatmaps are produced. Finally, the transformations are reversed and the two outputs merged.}
    \label{fig:siamese-training}
    \vspace{-10px}
\end{figure}

Largely, the face alignment training procedure has remained unchanged since the very first deep learning methods of~\cite{bulat2017far,zhu2016face}. Herein, we propose to deviate from this paradigm adopting a Siamese-based training, where two different random augmentations of the same image are passed through the network, producing in the process a set of heatmaps. We then revert the transformation of each of these heatmaps and combine them via element-wise summation.

The advantages of this training process are twofold: Firstly, convolutional networks are not invariant under arbitrary affine transformations, and, as such, relatively small variances in the input space can result in large differences in the output. Therefore, by optimizing jointly and combining the two predictions we can improve the consistency of the predictions.

Secondly, while previously the 2D Gaussians were always centered around an integer pixel location due to the quantization of the coordinates via rounding, the newly proposed heatmap generation can have the center in-between (\ie on a sub-pixel). As such, to avoid small sub-pixel inconsistencies and misalignment introduced by the data augmentation process we adopt the above-mentioned Siamese based training. Our approach, depicted in Fig.~\ref{fig:siamese-training}, defines the output heatmaps $\hat{\pazocal{H}}$ as: 

\begin{equation}
    \tilde{\pazocal{H}} = T_0^{-1}(\Phi(T_0(\textbf{X}_i), \theta)) + T_1^{-1}(\Phi(T_1(\textbf{X}_i)), \theta),
\end{equation}
where $\Phi$ is the network for heatmap regression with parameters $\theta$. $T_0$ and $T_1$ are two random transformations applied on the input image $\textbf{X}_i$ and, $T_0^{-1}$ and $T_1^{-1}$ denote their inverse.
\vspace{-10px}
\section{Ablation studies}~\label{sec:ablation}
\vspace{-30px}

\subsection{Comparison with other landmarks localization losses}\label{ssec:comparison-loss}

Beyond comparisons with recently proposed methods for face alignment in Section~\ref{sec:sota} (e.g.~\cite{feng2018wing,wang2019adaptive,kumar2020luvli}), herein we compare our approach against a few additional baselines.

\begin{wraptable}[12]{r}{0.49\textwidth}
    \vspace{-0.3cm}
    \centering
    \begin{tabular}{cc}
        \toprule
        Method & $\text{NME}_{box}$ \\
        \midrule
         $\ell_2$ heatmap regression & 2.32\\
         coord-correction (static gt) & 2.27\\
         coord-correction (dynamic gt) & 2.30 \\
         Global soft-argmax & 3.19 \\
         \midrule
         Local soft-argmax (Ours) & 2.04\\
    \bottomrule
    \end{tabular}
        \caption{Comparison between various losses baselines on 300W test set.}
    \label{tab:ablation-loss}
\end{wraptable}

\noindent\textbf{Heatmap prediction with coordinate correction:} In DeepCut~\cite{pishchulin2016deepcut}, for human pose estimation, the authors propose to add a coordinate refinement layer that predicts a $(\Delta\hat{y}_k^{[1]},\Delta\hat{y}_k^{[2]})$ displacement that is then added to the integer predictions generated by the heatmaps. To implement this, we added a global pooling operation followed by a fully connected layer and then trained it jointly using an $\ell_2$ loss. We attempted 2 different variants: one where the $(\Delta\hat{y}_k^{[1]},\Delta\hat{y}_k^{[2]})$ is constructed by measuring the heatmap encoding errors and the other is dynamically constructed at runtime by measuring the error between the heatmap prediction and the ground truth. As Table~\ref{tab:ablation-loss} shows, these learned corrections offer minimal improvements on top of the standard heatmap regression loss and are noticeably worse than the accuracy scored by the proposed method. This shows that predicting sub-pixel errors using a second branch is less effective than constructing better heatmaps from the first place.

\noindent\textbf{Global soft-argmax:} In~\cite{luvizon20182d}, the authors propose to to predict the locations of the points of interest on the human body by estimating their position using a global soft-argmax as a differentiable alternative to taking the argmax. From a first glance this is akin to the idea proposed in this work: local soft-argmax. However, applying soft-argmax globally leads to semantically unstructured outputs~\cite{luvizon20182d} that hurt the performance. Even adding a Gaussian prior is insufficient for achieving high accuracy on face alignment. As the results from Table~\ref{tab:ablation-loss} conclusively show, our simple improvement, namely the proposed local soft-argmax is the key idea for obtaining highly accurate results.

\begin{figure*}
    \centering
     \begin{subfigure}{0.235\textwidth}
         \includegraphics[width=\textwidth]{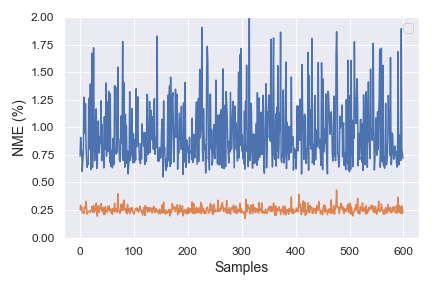}
         \caption{300W}
     \end{subfigure}
     ~
    \begin{subfigure}{0.235\textwidth}
         \includegraphics[width=\textwidth]{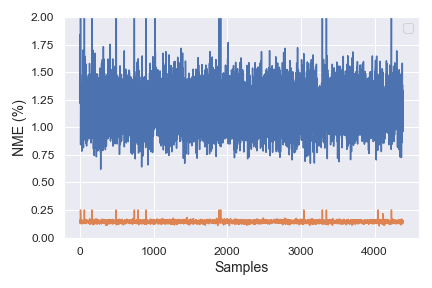}
         \caption{AFLW}
     \end{subfigure}
     ~
     \begin{subfigure}{0.235\textwidth}
         \includegraphics[width=\textwidth]{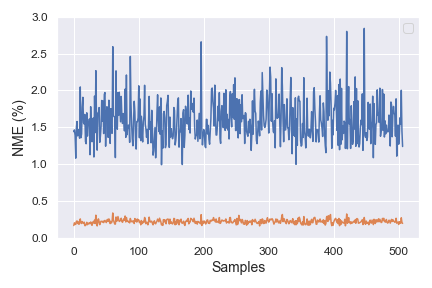}
         \caption{COFW}
     \end{subfigure}
     ~
    \begin{subfigure}{0.235\textwidth}
         \includegraphics[width=\textwidth]{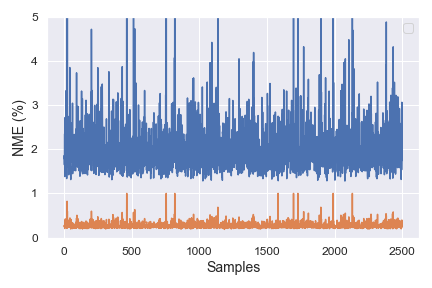}
         \caption{WFLW}
     \end{subfigure}
     ~     
    \caption{NME after encoding and then decoding of the ground truth heatmaps for various datasets using our proposed approach (orange) and the standard one~\cite{bulat2017far} (blue). Notice that our approach significantly reduces the error rate across all samples from the datasets. }
    \label{fig:error_rates}
    \vspace*{-15px}
\end{figure*}

\begin{wraptable}[9]{r}{0.4\textwidth}
\vspace{-0.3cm}
    \centering
    \resizebox{0.4\columnwidth}{!}{%
    \begin{tabular}{cc}
    \toprule
       Method & $\text{NME}_{ic}$ (\%) \\
       \midrule
       Baseline~\cite{bulat2017far}  &  4.20\\
       + proposed hm  & 3.90\\
       + proposed hm (w/o~\ref{ssec:heatmap-decoding}) & 4.00\\
       + siamese training & 3.72\\
       \bottomrule
    \end{tabular}
    }
    \caption{Effect of the proposed components on the WFLW dataset.}
    \label{tab:effect_of_components}
\end{wraptable}

\vspace{-10px}
\subsection{Effect of method's components}\label{ssec:effect-component}

Herein, we explore the impact of each our method's component on the overall performance of the network. As the results from Table~\ref{tab:effect_of_components} show, starting from the baseline introduced in~\cite{bulat2017far}, the addition of the proposed heatmap encoding and decoding process significantly improves the accuracy. If we analyze this result in tandem with Fig.~\ref{fig:error_rates} it becomes apparent what is the source of these gains: In particular, Fig.~\ref{fig:error_rates} shows the heatmap encoding and decoding process of the baseline method~\cite{bulat2017far} as well as of our method using directly the ground truth landmarks (i.e. these are not network's predictions). As shown in Fig.~\ref{fig:error_rates}, simply encoding and decoding the heatmaps corresponding to the ground truth alone induces high NME for~\cite{bulat2017far}. While the training procedure is able to compensate this, these inaccuracies representations hinder the learning process. Furthermore, due to the sub-pixel errors introduced, the performance in the high accuracy regime of the cumulative error curve degrades.

The rest of the gains are achieved by switching to the proposed Siamese training that reduces the discrepancies between multiple views of the same image while also reducing potential sub-pixel displacements that may occur between the image and the heatmaps.
\vspace{-10px}
\subsection{Local window size}\label{ssec:windows-size}

In this section, we analyze the relation between the local soft-argmax window size and the model's accuracy. As the results from Table~\ref{tab:window_size} show, the optimal window has a size of $5\times5$px, which corresponds to the size of the generated gaussian (i.e., most of the non-zero values will be contained within this window). Furthermore, as the window size increases the amount of noise and background pixels also increases and hence the accuracy decreases. The same value is used across all datasets. Note, that explicitly using the local window loss during training doesn't improve the performance further which suggest that the pixel-wise loss alone is sufficient, if the encoding process is accurate.  

\begin{wraptable}[7]{r}{0.49\textwidth}
\vspace{-0.3cm}
    \centering
    \begin{tabular}{c|cccc}
    \toprule
          & none & $3\times 3$ & $5\times 5$ & $7 \times 7$  \\
         \midrule
       $\text{NME}_{box}$ & 2.21 & 2.06 & 2.04 & 2.07 \\
     \bottomrule
    \end{tabular}
        \caption{Effect of window size on the 300W test set.}
    \label{tab:window_size}
\end{wraptable}

\vspace{-0.3cm}
\section{Experimental setup}

\textbf{Datasets:}
We preformed extensive evaluations to quantify the effectiveness of the proposed method. We trained and/or tested our method on the following datasets: 300W~\cite{sagonas2013300} (constructed in~\cite{sagonas2013300} using images from LFPW~\cite{belhumeur2013localizing}, AFW~\cite{zhu2012face}, HELEN~\cite{le2012interactive} and iBUG~\cite{sagonas2013semi}), 300W-LP~\cite{zhu2016face}, Menpo~\cite{zafeiriou2017menpo}, COFW-29~\cite{burgos2013robust}, COFW-68~\cite{ghiasi2015occlusion}, AFLW~\cite{kostinger2011annotated}, WFLW~\cite{wayne2018lab} and 300VW~\cite{shen2015first}. For a detailed description of each dataset see supplementary material.
\newline\newline
\noindent\textbf{Metrics:}
Depending on the evaluation protocol of each dataset we used one or more of the following metrics:

\noindent\textbf{Normalized Mean Error (NME)} that measures the point-to-point normalized Euclidean distance. Depending on the testing protocol, the NME $type$ will vary. In this paper, we distinguish between the following types:  $d_{ic}$ -- computed as the inter-occular distance~\cite{sagonas2013300}, $d_{box}$ -- computed as the geometric mean of the ground truth bounding box~\cite{bulat2017far} $d = \sqrt(w_{bbox} \cdot h_{bbox})$, and finally $d_{diag}$ -- defined as the diagonal of the bounding box.

\noindent\textbf{Area Under the Curve(AUC):} The AUC is computed by measuring the area under the curve up to a given user defined cut-off threshold of the cumulative error curve.

\noindent\textbf{Failure Rate (FR):} The failure rate is defined as the percentage of images the NME of which is bigger than a given (large) threshold.

\subsection{Training details}\label{ssec:training-details}
For training the models used throughout this paper we largely followed the common best practices from literature. Mainly, during training we applied the following augmentation techniques: Random rotation (between $\pm 30^{o}$), image flipping and color($0.6$, $1.4$) and scale jittering (between $0.85$ and $1.15$). The models where trained for 50 epochs using a step scheduler that dropped the learning rate at epoch 25 and 40 starting from a starting learning rate of $0.0001$. Finally, we used Adam~\cite{kingma2014adam} for optimization. 
The predicted heatmaps were at a resolution of $64\times64$px, i.e. $4\times$ smaller than the input images which were resized to $256\times 256$ pixels with the face size being approximately equal to $220\times220$px. The network was optimized using an $\ell_2$ pixel-wise loss. For the heatmap decoding process, the temperature of the soft-argmax $\tau$ was set to 10 for all datasets, however slightly higher values perform similarly. Values that are too small or high would ignore and respectively overly emphasise the pixels found around the coordinates of the max.
All the experiments were implemented using PyTorch~\cite{NEURIPS2019_9015} and Kornia~\cite{riba2020kornia}.

\noindent\textbf{Network architecture:} All models trained throughout this work, unless otherwise specified, follow a 2-stack Hourglass based architecture with a width of 256 channels, operating at a resolution of $256\times256$px as introduced in~\cite{bulat2017far}. Inside the hourglass, the features are rescaled down-to $4\times4$px and then upsampled back, with skip connection linking features found at the same resolution. The network is constructed used the building block from~\cite{bulat2017binarized} as in~\cite{bulat2017far}.  For more details regarding the network structure see~\cite{newell2016stacked,bulat2017far}.

\vspace{-0.3cm}
\section{Comparison against state-of-the-art}\label{sec:sota}

Herein, we compare against the current state-of-the-art face alignment methods across a plethora of datasets. Throughout this section the best result is marked in table with bold and red while the second best with bold and blue color. The important finding of this section is by means of two simple improvements: (a) improving the heatmap encoding and decoding process and, (b) including the Siamese training, we managed to obtain results which are significantly better than all recent prior work, setting in this way a new state-of-the-art.

\noindent \textbf{Comparison on WFLW:} On WFLW, and following their evaluation protocol, we report results in terms of $\text{NME}_{ic}$, $\text{AUC}^{10}_{ic}$ and $\text{FR}^{10}_{ic}$. As the results from Table~\ref{tab:sota_wflw} show, our method improves the previous best results of~\cite{kumar2020luvli} by more than $0.5\%$ for $\text{NME}_{ic}$ and 5\%  in terms of $\text{AUC}^{10}_{ic}$ almost halving the error rate. This shows that our method  offers improvements in the high accuracy regime while also reducing the overall failure ratio for difficult images. 

\noindent \textbf{Comparison on AFLW:} Following~\cite{kumar2020luvli}, we report results in terms of  $\text{NME}_{diag}$, $\text{NME}_{box}$ and $\text{AUC}^{7}_{box}$. As the results from Table~\ref{tab:sota_aflw19} show, we improve across all metrics on top of the current best result even on this nearly saturated dataset.

\begin{table}[!htbp]
	\centering
	\begin{subtable}{.48\textwidth}
	\resizebox{0.99\textwidth}{!}{
		\begin{tabular}{cccc}
			\toprule
			Method & $\text{NME}_{ic}$(\%) & $\text{AUC}^{10}_{ic}$ & $\text{FR}^{10}_{ic}$ (\%) \\
			\midrule
            Wing~\cite{feng2018wing} & 5.11 & 0.554 & 6.00 \\
            MHHN~\cite{wan2020robust} & 4.77 & - & \\
            DeCaFa~\cite{dapogny2019decafa} & 4.62 & 0.563 & 4.84  \\
            AVS~\cite{qian2019aggregation} & 4.39 & \secondb{0.591} & 4.08  \\
            AWing~\cite{wang2019adaptive} & 4.36 & 0.572 & \secondb{2.84}  \\
            LUVLi~\cite{kumar2020luvli} & 4.37 & 0.577 & 3.12 \\
            GCN~\cite{li2020structured} & \secondb{4.21} & 0.589 & 3.04 \\
            Ours & \best{3.72} & \best{0.631} & \best{1.55} \\ 
			\bottomrule
		\end{tabular}
		}
			\caption{Comparison against the state-of-the-art on WFLW in terms of $\text{NME}_{inter-ocular}$, $\text{AUC}^{10}_{ic}$ and $\text{FR}^{10}_{ic}$.}\label{tab:sota_wflw}
	\end{subtable}
		\begin{subtable}{.48\textwidth}
		\resizebox{0.99\textwidth}{!}{
		\begin{tabular}{cccc}
			\toprule
			& Common & Challenge & Full \\
			\midrule
            Teacher~\cite{dong2019teacher} & 2.91 & 5.91 & 3.49 \\
            DU-Net~\cite{tang2019towards} & 2.97 & 5.53 & 3.47 \\
            DeCaFa~\cite{dapogny2019decafa} & 2.93 & 5.26 & 3.39 \\
            HR-Net~\cite{sun2019high} & 2.87 & 5.15 & 3.32 \\
            HG-HSLE~\cite{zou2019learning} & 2.85 & 5.03 & 3.28 \\
            Awing~\cite{wang2019adaptive} & \secondb{2.72} & \secondb{4.52} & \secondb{3.07} \\
            LUVLi~\cite{kumar2020luvli} & 2.76 & 5.16 & 3.23 \\
            Ours & \best{2.61} & \best{4.13} &  \best{2.94} \\
			\bottomrule
		\end{tabular}
		}
				\caption{Comparison against state-of-the-art on the 300W Common, Challenge and Full datasets (\ie Split II) in terms of $\text{NME}_{inter-occular}$}\label{tab:sota_300w_valid}
	\centering
	\end{subtable}
	\caption{Results on WFLW (a) and 300W (b) datasets.}
\end{table}

\begin{wraptable}[10]{r}{0.49\textwidth}
\vspace{-0.3cm}
	\centering
		\begin{tabular}{ccc}
			\toprule
			Method & $\text{NME}_{ic}$(\%) &  $\text{FR}^{10}_{ic}$ (\%) \\
			\midrule
            Wing~\cite{feng2018wing} & 5.07 & 3.16 \\
            LAB  (w/B)~\cite{wu2018look} & 3.92 & 0.39 \\
            HR-Net~\cite{sun2019high} & \secondb{3.45} & \secondb{0.19} \\
            Ours & \best{3.02} & \best{0.0} \\ 
			\bottomrule
		\end{tabular}
			\caption{Comparison on COFW-29. Results for other methods taken from~\cite{sun2019high}.}\label{tab:sota_cofw29}
\end{wraptable}

\noindent \textbf{Comparison on 300W:} Following the protocol described in~\cite{sagonas2013300} and~\cite{bulat2017far}, we report results in terms of $\text{NME}_{inter-occular}$ for \textit{Split I} and of $\text{AUC}^7_{box}$ and $\text{NME}_{box}$ for \textit{split II}. Note that due to the overlap between the splits we train two separate models, one on the data from the first split and another on the data from the other split evaluating the models accordingly. Following~\cite{bulat2017far,kumar2020luvli} the model evaluated on the test set was pretrained on 300W-LP dataset. As the results from Table~\ref{tab:sota_300w_valid} show, our approach offers consistent improvements across both subsets (\ie \textit{Common} and \textit{Challenge}), with particularly higher gains on the later. Similar results can be observer in Table~\ref{tab:sota_300w_menpo_cofw68} for \textit{Split II}.

\noindent \textbf{Comparison on COFW:} On the COFW dataset we evaluate on both the 29-point (see Table~\ref{tab:sota_cofw29}) and 68-point configuration (see Table~\ref{tab:sota_300w_menpo_cofw68}) in terms of $\text{NME}_{ic}$(\%) and $\text{FR}^{10}_{ic}$ for the 29-point configuration and $\text{NME}_{box}$, $\text{AUC}^7_{box}$ for the other one. As the results from Tables~\ref{tab:sota_300w_menpo_cofw68} and~\ref{tab:sota_cofw29} show, our method sets a new state-of-the-art, reducing the failure rate to 0.0.

\noindent \textbf{Comparison on Menpo:} Following~\cite{kumar2020luvli} we evaluate on the frontal sub-set of the Menpo dataset. As Table~\ref{tab:sota_300w_menpo_cofw68} shows, our method sets a new state-of-the-art result.

\noindent \textbf{Comparison on 300VW:} Unlike the previous datasets that focus on face alignment for static images, 300VW is a video face tracking dataset. Following~\cite{shen2015first}, we report results in terms of $\text{AUC}_{ic}@0.08$ on the most challenging partition of the test set (C). As the results from Table~\ref{tab:sota_300vw} show, despite not exploiting any temporal information and running our method on a frame-by-frame basis, we set a new state-of-the-art, outperforming previous tracking methods trained  such as~\cite{sanchez2017functional} and~\cite{haris2017synergy}. Similar results can be observed when evaluating on all 68 points in Table~\ref{tab:sota_300vw_nme}.

\begin{table}
	\centering
	\resizebox{0.6\textwidth}{!}{
		\begin{tabular}{ccccc}
			\toprule
			\multirow{2}{*}{Method} &  \multicolumn{2}{c}{$\text{NME}_{diag}$} & $\text{NME}_{box}$ & $\text{AUC}^{7}_{box}$ \\
			\cline{2-5}
			& Full & Frontal & Full & Full \\
			\midrule
            SAN~\cite{dong2018style} & 1.91 & 1.85 & 4.04 & 54.0 \\
            DSNR~\cite{miao2018direct} & 1.85 & 1.62 & - & - \\
            LAB (w/o B)~\cite{wu2018look} & 1.85 & 1.62 & - & - \\
            HR-Net~\cite{sun2019high} & 1.57 & 1.46 & - & - \\
            Wing~\cite{feng2018wing} & - & - & 3.56 & 53.5 \\
            KDN~\cite{chen2019face} & - & - & 2.80 & 60.3 \\
            LUVLi~\cite{kumar2020luvli} & 1.39 & \secondb{1.19} & \secondb{2.28} & \secondb{68.0} \\
            MHHN~\cite{wan2020robust} & \secondb{1.38} & \secondb{1.19} & - & - \\
            Ours & \best{1.31} & \best{1.12} & \best{2.14} & \best{70.0} \\
			\bottomrule
		\end{tabular}
		}
			\caption{Comparison against the state-of-the-art on the AFLW-19 dataset.}\label{tab:sota_aflw19}
			\vspace{-10px}
\end{table}

\begin{table*}
	\centering
	\resizebox{0.8\textwidth}{!}{
		\begin{tabular}{ccccccc}
			\toprule
			\multirow{2}{*}{Method} &  \multicolumn{3}{c}{$\text{NME}_{box}$} & \multicolumn{3}{c}{$\text{AUC}^7_{box}$} \\
			\cline{2-4} \cline{5-7}
			& 300-W & Menpo & COFW-68 & 300-W & Menpo & COFW-68  \\
			\midrule
            SAN~\cite{dong2018style} & 2.86 & 2.95 & 3.50 & 59.7 & 61.9 & 51.9 \\
            FAN~\cite{bulat2017far} & 2.32 & 2.16 & 2.95 & 66.5 & 69.0 & 57.5 \\
            Softlabel~\cite{chen2019face} & 2.32 & 2.27 & 2.92 & 66.6 & 67.4 & 57.9 \\
            KDN~\cite{chen2019face} & 2.21 & \secondb{2.01} & 2.73 & 68.3 & 71.1 & 60.1 \\
            LUVLi~\cite{kumar2020luvli} & \secondb{2.10} & 2.04 & \secondb{2.57} & \secondb{70.2} & \secondb{71.9} & \secondb{63.4} \\
            Ours & \best{2.04} & \best{1.95} & \best{2.47} & \best{71.1} & \best{73.0} & \best{64.9} \\
			\bottomrule
		\end{tabular}
		}
		\caption{Comparison against the state-of-the-art on the 300W Test (\ie Split I), Menpo 2D Frontal and COFW-68 datasets in terms of $\text{NME}_{box}$ and $\text{AUC}^7_{box}$.}\label{tab:sota_300w_menpo_cofw68}
\end{table*}

\begin{table*}
	\centering
	\resizebox{0.99\textwidth}{!}{
		\begin{tabular}{ccccccc}
			\toprule
			Method &  \textbf{Ours} & DGM~\cite{haris2017synergy} & CPM+SRB+PAM~\cite{dong2018supervision} & iCCR~\cite{sanchez2017functional} & \cite{yang2015facial} & \cite{xiao2015facial} \\
			\midrule
            $\text{AUC}_{ic}@0.08$ & \best{60.10} & 59.38 & \secondb{59.39} & 51.41 & 49.96 & 48.65 \\
			\bottomrule
		\end{tabular}
		}
			\caption{Comparison against the state-of-the-art on the 300-VW dataset -- category C, in terms of $\text{AUC}_{ic}@0.08$ evaluated on the 49 inner points.}\label{tab:sota_300vw}
	\vspace*{-0.4cm}
\end{table*}

\begin{table*}
	\centering
		\begin{tabular}{ccccccc}
			\toprule
			Method &  \textbf{Ours} & FHR+STA~\cite{tai2019towards} & TSTN~\cite{liu2017two} & TCDCN~\cite{zhang2015learning} & CFSS~\cite{zhu2015cfss}\\
			\midrule
            $\text{NME}_{ic}$ & \best{5.84} & \secondb{5.98} & 12.80 & 15.0  &13.70 \\
			\bottomrule
		\end{tabular}
			\caption{Comparison against the state-of-the-art on the 300-VW dataset -- category C (i.e., scenario 3), in terms of $\text{NME}_{ic}$ evaluated on all 68 points. Results for other methods taken from~\cite{tai2019towards}.}\label{tab:sota_300vw_nme}
	\vspace*{-0.4cm}
\end{table*}

\begin{figure*}[htb!]
    \centering
    \begin{tcbraster}[raster columns=5, enhanced, blankest, raster row skip=1mm, raster column skip=1mm]
    \tcbincludegraphics{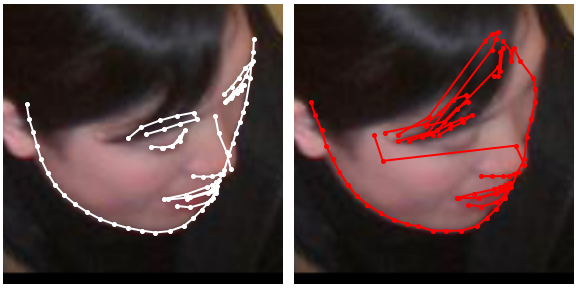}
    \tcbincludegraphics{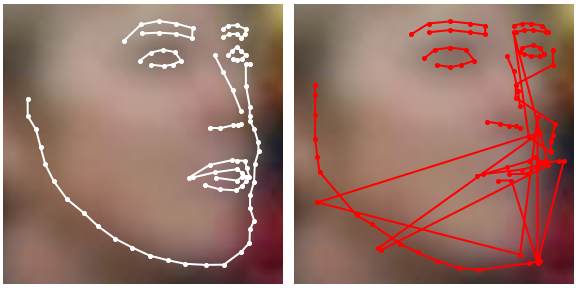}
    \tcbincludegraphics{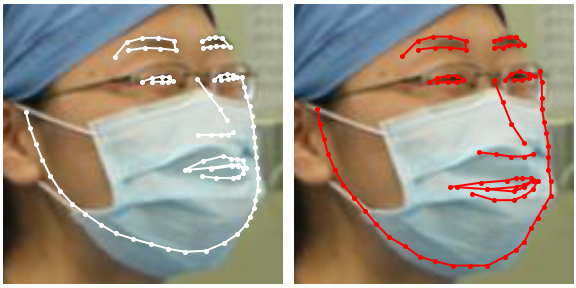}
    \tcbincludegraphics{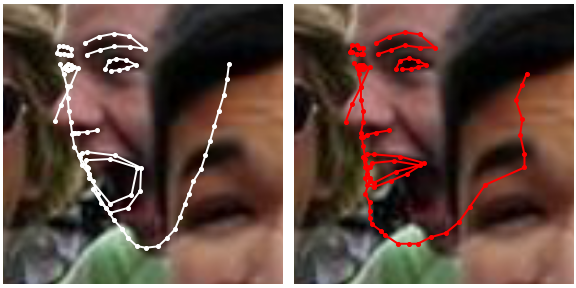}
    \tcbincludegraphics{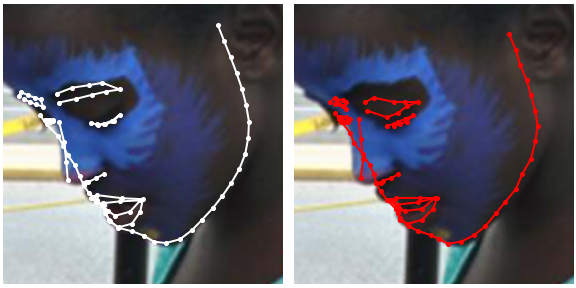}
    
    \tcbincludegraphics{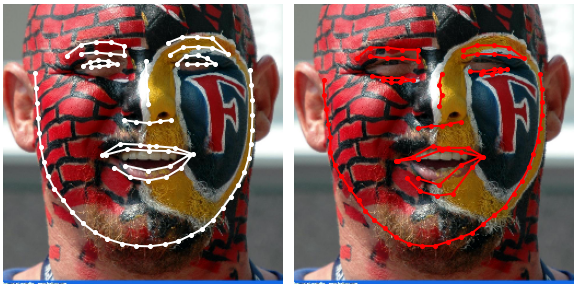}
    \tcbincludegraphics{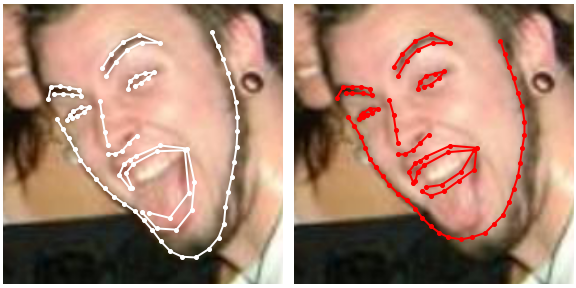}
    \tcbincludegraphics{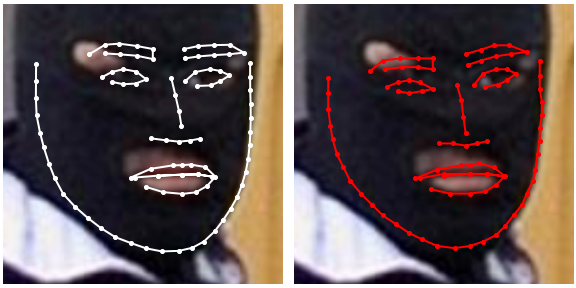}
    \tcbincludegraphics{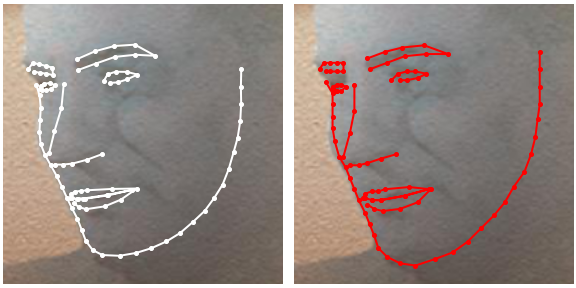}
    \tcbincludegraphics{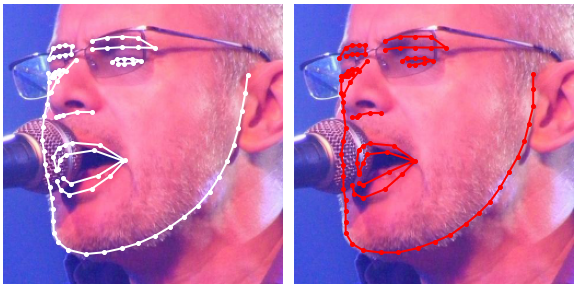}
    \end{tcbraster}
    \caption{Qualitative results. Landmarks shown in \underline{white} are produced by our method, while the ones in \underline{red} by the  state-of-the-art approach of~\cite{bulat2017far}. Thanks to the proposed heatmap encoding and decoding, our method is able to provide much more accurate results.  Best viewed zoomed in, in electronic format.}\label{fig:examples}
\end{figure*}

\begin{figure*}[htb!]
    \centering
    \begin{tcbraster}[raster columns=7, enhanced, blankest, raster row skip=1mm, raster column skip=1mm]
    \tcbincludegraphics[graphics options={trim=0 0 10.5cm 0, clip}]{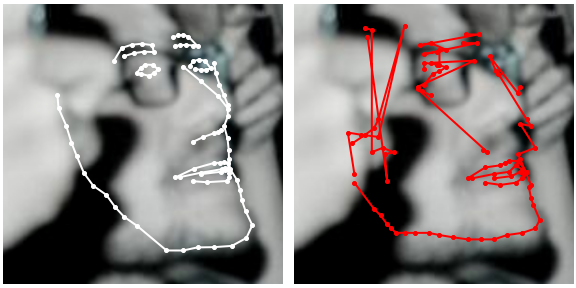}
    \tcbincludegraphics[graphics options={trim=0 0 10.5cm 0, clip}]{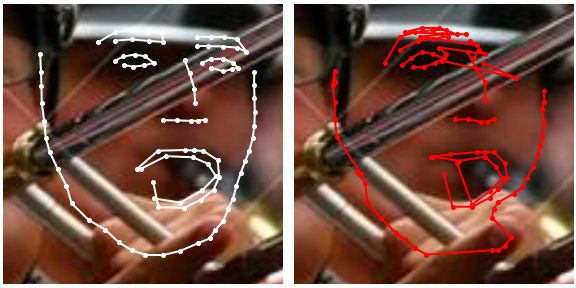}
    \tcbincludegraphics[graphics options={trim=0 0 10.5cm 0, clip}]{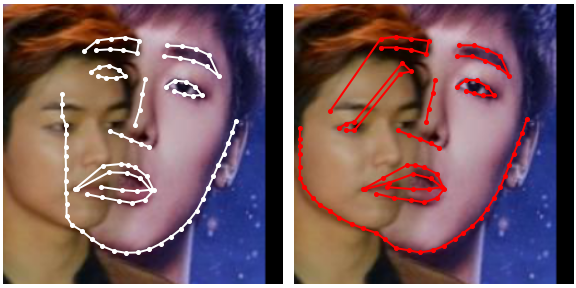}
    \tcbincludegraphics[graphics options={trim=0 0 10.5cm 0, clip}]{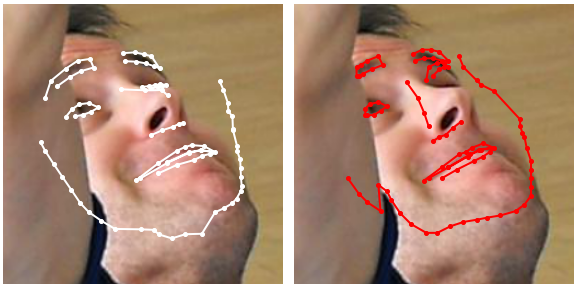}
    \tcbincludegraphics[graphics options={trim=0 0 10.5cm 0, clip}]{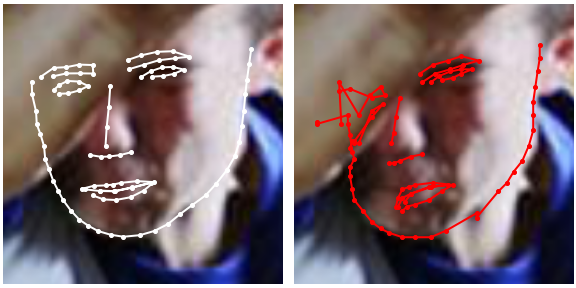}
    \tcbincludegraphics[graphics options={trim=0 0 10.5cm 0, clip}]{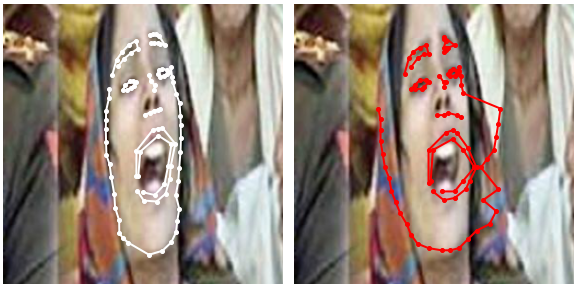}
    \tcbincludegraphics[graphics options={trim=0 0 10.5cm 0, clip}]{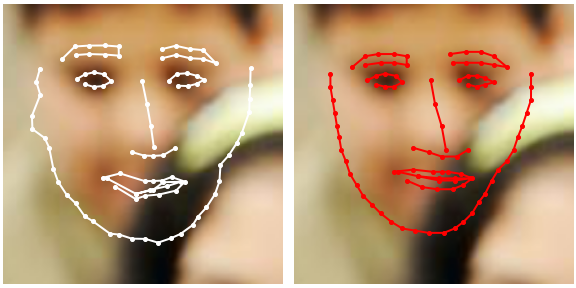}
    \end{tcbraster}
    \caption{Examples of failure cases. Most of the failure cases include combinations of low resolution images with extreme poses (1st and 4th image), perspective distortions (5th image) or overlapping faces (3rd image). }\label{fig:examples-failure}
\end{figure*}

\section{Conclusions}
\vspace{-5px}
We presented simple yet effective improvements to standard methods for face alignment which are shown to dramatically increase the accuracy on all benchmarks considered without introducing sophisticated changes to existing architectures and loss functions. The proposed improvements concern a fairly unexplored topic in face alignment that of the heatmap encoding and decoding process. We showed that the proposed continuous heatmap regression provides a significantly improved approach for the encoding/decoding process. Moreover, we showed that further improvements can be obtained by considering a simple Siamese training procedure that enforces output spatial consistency of geometrically transformed images. We hope that these improvements will be incorporated in future research while it is not unlikely that many existing methods will also benefit by them.
\bibliography{refs}

\appendix

\section{Datasets}

In this paper we conducton experiments on the following datasets:

\noindent\textbf{300W:} 300W~\cite{sagonas2013300} is a 2D face alignment dataset constructed by concatenating and then manually re-anotating with 68 points the images from LFPW~\cite{belhumeur2013localizing}, AFW~\cite{zhu2012face}, HELEN~\cite{le2012interactive} and iBUG~\cite{sagonas2013semi}. There are two commonly used train/test splits. Split I: uses 3837 images for training and 600 for testing and Split II that uses 3148 facial images for training and 689 for testing. The later testset comprises of two subsets: common and challenge.  Most of the images present in the dataset contain faces found in frontal or near-frontal poses.

\noindent\textbf{300W-LP:} 300W-LP~\cite{zhu2016face} is a synthetically generated dataset formed by warping into large poses the images from the 300W dataset. This dataset contains 61,125 pre-warped images and is used for training alone.

\noindent\textbf{Menpo:} Menpo~\cite{zafeiriou2017menpo} is a 2D face alignment dataset that annotates the images using 2 different configurations depending on the pose of the faces. The near frontal facial images are annotated using the same 68 points configuration used for 300W, while the rest using 39 points. In this work, we trained and evaluated on the 68-point configuration.

\noindent\textbf{COFW:} The Caltech Occluded Faces in the Wild (COFW)~\cite{burgos2013robust} dataset contains 1,345 training and 507 testing facial images captured in real world scenarios and annotated using 29 points. The images were later on re-annotated in~\cite{ghiasi2015occlusion} using the same 68-point configuration as in 300W.

\noindent\textbf{AFLW:} The Annotated Facial Landmarks in the Wild (AFLW)~\cite{kostinger2011annotated} dataset consists of 20,000 training images and 4386 testing images, out of which 1314 are part of the \textit{Frontal} subset. All images are annotated using a 19 point configuration.

\noindent\textbf{WFLW:} Wider Facial Landmarks in-the-wild (WFLW)~\cite{wayne2018lab} consists of 10,000 images, out of which 7,500 are used for training while the rest are reserved for testing. All images are annotated using a 98 point configuration. In addition to landmarks, the dataset is also annotated with a set of attributes.

\noindent\textbf{300VW:} 300VW~\cite{shen2015first} is a large scale video face alignment dataset consisting of 218,594 frames distributed across 114 videos, out of which 50 are reserved for training while the rest for testing. The test set is further split into 3 different categories (A, B an C) with C  being the most challenging one. We note that due to the semi-supervised annotation procedure some images have erroneous labels.

\section{Metrics}

Depending on the dataset, the following metrics were used throughout this work:

\noindent\textbf{Normalized Mean Error (NME)}. The point-to-point normalized Euclidean distance is the most widely used metric for evaluating the accuracy of a face alignment method and is defined as:
    $\text{NME}_{type} (\%)  = \frac{1}{N} \sum_{k}^N \mathbf{v}_{k} \frac{\mathbf{y_k} - \mathbf{\hat{y}_k}}{d_{type}} \times 100,$
where $\mathbf{y}_k$ denotes the ground truth landmarks for the $k$-th face, $\mathbf{\hat{y}_k}$ its corresponding predictions and $d_{type}$ is the reference distance by which the points are normalized. $\mathbf{v}_k$ is a visibility binary vector, with values 1 at the landmarks where the ground truth is provided and 0 everywhere else.

Depending on the testing protocol, the NME $type$ (\ie how it's computed and defined) will vary. In this paper, we distinguish between the following types:  $d_{ic}$ -- computed as the inter-occular distance~\cite{sagonas2013300}, $d_{box}$ -- computed as the geometric mean of the ground truth bounding box~\cite{bulat2017far} $d = \sqrt(w_{bbox} \cdot h_{bbox})$, and finally $d_{diag}$ -- defined as the diagonal of the bounding box.

\noindent\textbf{Area Under the Curve(AUC):} The AUC is computed by measuring the area under the curve up to a given user defined cut-off threshold of the cumulative error curve. Compared with NME that simple takes the average, this metric is less prone to outliers.

\noindent\textbf{Failure Rate (FR):} The failure rate is defined as the percentage of images the NME of which is bigger than a given (large) threshold.

\section{Additional comparisons with state-of-the-art}

In addition to the comparisons reported in the main paper here in we show how our method performs against an additional set of methods (Tables~\ref{tab:sota_wflw},~\ref{tab:sota_aflw19},~\ref{tab:sota_300w_valid},~\ref{tab:sota_300w_valid},~\ref{tab:sota_cofw29}).

\begin{table}[!htbp]
	\caption{Comparison against the state-of-the-art on WFLW in terms of $\text{NME}_{inter-ocular}$, $\text{AUC}^{10}_{ic}$ and $\text{FR}^{10}_{ic}$.}\label{tab:sota_wflw}
	\centering
		\begin{tabular}{cccc}
			\toprule
			Method & $\text{NME}_{ic}$(\%) & $\text{AUC}^{10}_{ic}$ & $\text{FR}^{10}_{ic}$ (\%) \\
			\midrule
			ESR~\cite{cao2014face} & 11.13 & 0.277 & 35.24 \\
			SDM~\cite{xiong2013supervised} & 10.29 & 0.300 & 29.40 \\
            CFSS~\cite{zhu2015cfss} & 9.07 & 0.366 & 20.56 \\
            DVLN~\cite{wu2017leveraging} & 6.08 & 0.456 & 10.84 \\
            LAB (w/B)~\cite{wu2018look} & 5.27 & 0.532 & 7.56 \\
            Wing~\cite{feng2018wing} & 5.11 & 0.554 & 6.00 \\
            MHHN~\cite{wan2020robust} & 4.77 & - & - \\
            DeCaFa~\cite{dapogny2019decafa} & 4.62 & 0.563 & 4.84  \\
            AVS~\cite{qian2019aggregation} & 4.39 & \secondb{0.591} & 4.08  \\
            AWing~\cite{wang2019adaptive} & 4.36 & 0.572 & \secondb{2.84}  \\
            LUVLi~\cite{kumar2020luvli} & 4.37 & 0.577 & 3.12 \\
            GCN~\cite{li2020structured} & \secondb{4.21} & 0.589 & 3.04 \\
            Ours & \best{3.72} & \best{0.631} & \best{1.55} \\ 
			\bottomrule
		\end{tabular}
\end{table}

\begin{table}[!htbp]
	\caption{Comparison against the state-of-the-art on the AFLW-19 dataset.}\label{tab:sota_aflw19}
	\centering
		\begin{tabular}{ccccc}
			\toprule
			\multirow{2}{*}{Method} &  \multicolumn{2}{c}{$\text{NME}_{diag}$} & $\text{NME}_{box}$ & $\text{AUC}^{7}_{box}$ \\
			\cline{2-5}
			& Full & Frontal & Full & Full \\
			\midrule
			RCPR~\cite{burgos2013robust} & 3.73 & 2.87 & - & - \\
            CFSS~\cite{zhu2015cfss} & 3.92 & 2.68 & - & - \\
            CCL~\cite{zhu2016unconstrained} & 2.72 & 2.17 & - & - \\
            DAC-CSR~\cite{feng2017dynamic} & 2.27 & 1.81 & - & - \\
            LLL~\cite{robinson2019laplace} & 1.97 & - & - & - \\
            CPM+SRB~\cite{dong2018supervision} & 2.14 & - & - & - \\
            SAN~\cite{dong2018style} & 1.91 & 1.85 & 4.04 & 54.0 \\
            DSNR~\cite{miao2018direct} & 1.85 & 1.62 & - & - \\
            LAB (w/o B)~\cite{wu2018look} & 1.85 & 1.62 & - & - \\
            HR-Net~\cite{sun2019high} & 1.57 & 1.46 & - & - \\
            Wing~\cite{feng2018wing} & - & - & 3.56 & 53.5 \\
            KDN~\cite{chen2019face} & - & - & 2.80 & 60.3 \\
            LUVLi~\cite{kumar2020luvli} & 1.39 & \secondb{1.19} & \secondb{2.28} & \secondb{68.0} \\
            MHHN~\cite{wan2020robust} & \secondb{1.38} & \secondb{1.19} & - & - \\
            Ours & \best{1.31} & \best{1.12} & \best{2.14} & \best{70.0} \\
			\bottomrule
		\end{tabular}
\end{table}

\begin{table}[!htbp]
	\caption{Comparison against state-of-the-art on the 300W Common, Challenge and Full datasets (\ie Split II).}\label{tab:sota_300w_valid}
	\centering
		\begin{tabular}{cccc}
			\toprule
			\multirow{2}{*}{Method} &  \multicolumn{3}{c}{$\text{NME}_{inter-occular}$} \\
			\cline{2-4}
			& Common & Challenge & Full \\
			\midrule
			ODN~\cite{zhu2019robust} & 3.56 & 6.67 & 4.17 \\
			CPM+SRB~\cite{dong2018supervision} & 3.28 & 7.58 & 4.10 \\ 
            SAN~\cite{dong2018style} & 3.34 & 6.60 & 3.98 \\
            AVS~\cite{qian2019aggregation} & 3.21 & 6.49 & 3.86 \\
            DAN~\cite{kowalski2017deep} & 3.19 & 5.24 & 3.59 \\
            LAB (w/B)~\cite{wu2018look} & 2.98 & 5.19 & 3.49 \\
            Teacher~\cite{dong2019teacher} & 2.91 & 5.91 & 3.49 \\
            DU-Net~\cite{tang2019towards} & 2.97 & 5.53 & 3.47 \\
            DeCaFa~\cite{dapogny2019decafa} & 2.93 & 5.26 & 3.39 \\
            HR-Net~\cite{sun2019high} & 2.87 & 5.15 & 3.32 \\
            HG-HSLE~\cite{zou2019learning} & 2.85 & 5.03 & 3.28 \\
            Awing~\cite{wang2019adaptive} & \secondb{2.72} & \secondb{4.52} & \secondb{3.07} \\
            LUVLi~\cite{kumar2020luvli} & 2.76 & 5.16 & 3.23 \\
            Ours & \best{2.61} & \best{4.13} &  \best{2.94} \\
			\bottomrule
		\end{tabular}
\end{table}

\begin{table}[!htbp]
	\caption{Comparison against the state-of-the-art on the COFW-29 dataset.}\label{tab:sota_cofw29}
	\centering
		\begin{tabular}{ccc}
			\toprule
			Method & $\text{NME}_{ic}$(\%) &  $\text{FR}^{10}_{ic}$ (\%) \\
			\midrule
            Human & 5.60 & - \\
            ESR~\cite{cao2014face} & 11.20 & 36.0 \\
            RCPR~\cite{burgos2013robust} & 8.50 & 20.00 \\
            HPM & 7.59 & 13.00 \\
            CCR~\cite{feng2014random} & 7.03 & 10.90 \\
            DRDA~\cite{zhang2016occlusion} & 6.49 & 6.00 \\
            RAR~\cite{xiao2016robust} & 6.03 & 4.14 \\
            DAC-CSR~\cite{feng2017dynamic} & 6.03 & 4.73 \\
            LAB (w/o B)~\cite{wu2018look} & 5.58 & 2.76 \\
            Wing~\cite{feng2018wing} & 5.07 & 3.16 \\
            MHHN~\cite{wan2020robust} & 4.95 & 1.78 \\
            LAB  (w/B)~\cite{wu2018look} & 3.92 & 0.39 \\
            HR-Net~\cite{sun2019high} & \secondb{3.45} & \secondb{0.19} \\
            Ours & \best{3.02} & \best{0.0} \\ 
			\bottomrule
		\end{tabular}
\end{table}

\end{document}